\documentclass[letterpaper, 10 pt, conference]{ieeeconf}  

\IEEEoverridecommandlockouts                              

\usepackage[utf8]{inputenc} 
\usepackage[T1]{fontenc}    
\usepackage{url}            
\usepackage{booktabs}       
\usepackage{amsfonts}       
\usepackage{amsmath}
\usepackage{mathtools}

\usepackage{nicefrac}       
\usepackage{microtype}      
\usepackage{xcolor}         
\usepackage{graphicx}
\usepackage{placeins}
\usepackage{subcaption}
\usepackage{multirow}
\usepackage{siunitx}
\usepackage[ruled]{algorithm2e}
\usepackage{wrapfig}
\usepackage[hidelinks]{hyperref}
\usepackage{color}
\hypersetup{
    colorlinks,
    linkcolor={blue},
    citecolor={blue},
    urlcolor={blue}
}

\usepackage{babel}

\newtheorem{observation}{Observation}

\newcommand\mypara[1]{\vspace{3pt}\noindent\textbf{#1.}}

\usepackage[some]{background}
\backgroundsetup{%
  scale=1,       
  angle=0,       
  opacity=0.7,    
  color = {darkgray},  
  contents={\parbox{\textwidth}{\centering 
 \large This paper has been accepted for publication at the \\2023 IEEE International Conference on Robotics and Automation (ICRA) \\[25cm]}
}
}

\title{\Large \bf 
Training Efficient Controllers via Analytic Policy Gradient

}

\author{
Nina Wiedemann$^{* 1}$, Valentin W\"uest$^{* 2}$, 
Antonio Loquercio$^{1}$,
Matthias M\"uller$^{3}$,\\ 
Dario Floreano$^{2}$, 
and Davide Scaramuzza$^{1}$
\thanks{$^*$These authors contributed equally and are listed alphabetically. $^1$~Robotics and Perception Group, University of Zurich, Switzerland.
$^2$~Laboratory of Intelligent Systems, \'Ecole Polytechnique F\'ed\'erale de Lausanne (EPFL).
$^3$~Embodied AI Lab, Intel.
This work was supported by the Swiss National Science Foundation through the National Centre of Competence in Research (NCCR)%
}
}

\begin{document}

\maketitle
\thispagestyle{empty}
\pagestyle{empty}

\begin{abstract}

Control design for robotic systems is complex and often requires solving an optimization to follow a trajectory accurately.
Online optimization approaches like Model Predictive Control (MPC) have been shown to achieve great tracking performance, but require high computing power.
Conversely, learning-based offline optimization approaches, such as Reinforcement Learning (RL), allow fast and efficient execution on the robot but hardly match the accuracy of MPC in trajectory tracking tasks. In systems with limited compute, such as aerial vehicles, an accurate controller that is efficient at execution time is imperative.
We propose an Analytic Policy Gradient (APG) method to tackle this problem. APG exploits the availability of differentiable simulators by training a controller offline with gradient descent on the tracking error. 
We address training instabilities that frequently occur with APG through curriculum learning and experiment on a widely used controls benchmark, the CartPole, and two common aerial robots, a quadrotor and a fixed-wing drone.
Our proposed method outperforms both model-based and model-free RL methods in terms of tracking error.
Concurrently, it achieves similar performance to MPC while requiring more than an order of magnitude less computation time.
Our work provides insights into the potential of APG as a promising control method for robotics.

\end{abstract}


\section*{Supplementary material}
\noindent
Video of ICRA presentation: \url{https://youtu.be/HdfvsI126D8}\\
Source code: \url{https://github.com/lis-epfl/apg_trajectory_tracking}

\BgThispage{}

\section{Introduction}\label{sec:intro}



Control systems for robotics are becoming increasingly complex, 
as they are required to cope with diverse environments and tasks. 
As a response to challenges in deploying robots in real-world environments, learning-based control has gained importance and has started to replace classical control such as Model Predictive Control (MPC) or PID control~\cite{li2019fast, yu2017deep}.
While both (reinforcement) learning and MPC can be used for trajectory tracking by maximizing a reward or minimizing a cost with respect to the reference trajectory, 
%
they drastically differ in terms of computation of the optimization.
MPC requires no prior training and computes the optimization directly at runtime, resulting in large computational demands on the robot.
Subsequently, much work is invested into approximation algorithms or tailored hardware, since otherwise, the computation is not feasible to execute on the real platform~\cite{salzmann2022neural, wang2021efficient, abdolhosseini2013efficient}. This is especially relevant for flying vehicles where only limited computing hardware can be carried. 
In contrast, model-free RL is often applied by shifting the optimization to a training phase in simulation before deployment. Although RL was developed for solving long-term planning problems, it can be used for trajectory tracking by negatively rewarding the distance to a reference~\cite{song2021autonomous}.
At runtime, inference of the policy then allows fast execution on the robot.
However, RL treats dynamics as a black-box model and probes the environment to gain information about it.
As a result, it is prone to require a large amount of training data, which is particularly restrictive for complex and slow-to-simulate systems.
Finding a solution requiring little training data and permitting fast computation during run-time would be desirable.

Recent developments in the field of differentiable simulators have given rise to an alternative approach called Analytic Policy Gradient (APG)~\cite{freeman2021Brax}, which has the potential to fulfill these two requirements.
In APG, differentiable simulators allow analytical calculation of the reward gradients, enabling direct training of a policy with backpropagation.
By employing prior knowledge of the system in the form of the analytical gradients, APG was demonstrated to rely on one to four orders of magnitude less training data than RL to achieve similar performance~\cite{hu2019difftaichi, de2018end, innes2019differentiable, qiao2020scalable}.
Since APG pre-trains a control policy, like model-free RL, it also allows fast execution at runtime and is flexibly applicable to even high-dimensional inputs such as images.


Nevertheless, APG faces several challenges when employed for robotic systems.
Firstly, to train a policy with APG, it is often required to propagate gradients backwards over several time steps, to when the causing action was applied. This paradigm is termed backpropagation-through-time (BPTT) in APG and, similar to training RNNs, it is known to cause vanishing or exploding gradients~\cite{bengio1994Learning}.
Secondly, APG is prone to get trapped in local minima~\cite{freeman2021Brax}.
Both problems may prohibit training of a robust and accurate policy.
Current implementations have thus focused on MuJoCo environments~\cite{freeman2021Brax, gillen2022Leveraging} or fluid simulations~\cite{holl2020Learning}, whereas experiments on complex robotic systems are still rare.
Furthermore, APG is usually only compared to RL and not to other performant robotics control approaches.
Detailed experiments are therefore still necessary to determine how its performance compares to established optimization-based control techniques, such as MPC~\cite{gillen2022Leveraging}.

In this work, we demonstrate the potential of APG as an efficient and accurate controller for flying vehicles. With a receding-horizon training and a curriculum learning strategy, our APG controller is able to learn trajectory-tracking tasks offline for three diverse problems: (1)~balancing an inverted pendulum on a cart (standard task in RL known as CartPole), (2)~navigating towards a 3D target with a fixed-wing aircraft~\cite{oettershagen2014explicit}, and (3)~tracking a trajectory with a quadrotor. In contrast to previous work, we systematically compare APG to both learning-based approaches as well as classical control, and demonstrate its advantages in terms of runtime, tracking performance, sample efficiency for adaptation scenarios, and input flexibility, i.e. vision-based control. 
%

Our APG controller achieves similar or better tracking performance than commonly used model-based and model-free RL algorithms. It also achieves similar performance as MPC while reducing computation time by more than an order of magnitude.
This demonstrates the potential of APG and addresses the problem of unstable training,
enabling APG to be applied in challenging robot control tasks, where computation time for deployment and flexibility to sensor inputs is important.
\section{Background and Related Work}\label{sec:related}

We first introduce a framework describing the problem setting mathematically.
We situate previous works within this framework and highlight the differences to our approach.

\subsection{Problem setting}

In contrast to the traditional RL setting, we target fixed-horizon problems where a reference trajectory is given as $\{\nu_t\},\ \nu_t\in S$ is to be tracked accurately. Thus, we do not maximize a reward but minimize the Mean Squared Error between states and a reference trajectory.
We define the task as a discrete time, continuous-valued optimization problem:
\begin{equation}
    \min_{\pi}\\ J^*(\pi) = \min_{\pi} \mathbb{E}_{(s_t, v_t) \sim \rho(\pi)} [ C(f\big(s_t, \pi(s_t, \nu)\big), \nu_{t})],
    \label{eq:prob}
\end{equation}
where $C$ is a cost depending on a given reference state $\nu_{t} \in S$ and the robot state $s_t \in S$; $\rho(\pi)$ is the distribution of possible state-reference pairs  $\lbrace (s_0, \nu_0), \dots, (s_t, \nu_t)\rbrace$ induced by the policy $\pi$. The state at the next time step is given by the dynamics model $f$ with $s_{t+1} = f(s_t, \pi(s_t, \nu))$. 
The robot state is initialized to the provided reference ($s_0 = \nu_0$). 

As typical in the control literature~\cite{borrelli2017predictive}, we define total cost $C$ as the sum of the state cost ($\ell^p$-norm of the difference between the next state and the reference) and the control cost ($\ell^p$-norm of the actions) at time $t$, i.e. 
\begin{equation}
\begin{aligned}
& C(f\big(s_t, \pi(s_t, \nu)\big), \nu_{t}) 
\\ 
&= \Vert  f\big(s_t, \pi(s_t, \nu)\big) - \nu_{t+1} \Vert_p + 
 \Vert\pi(s_t, \nu)\Vert_p\ .
\end{aligned}
\end{equation}
%
%
%
%

\subsection{Model-free RL}

\begin{observation} $J^*({\pi})$ can be optimized directly with respect to $\pi$ by Monte-Carlo sampling to obtain empirical estimates of the gradients. This does not require $f$ to be differentiable. In turn, it requires extensive sampling, especially for long-term planning. This approach is used by both model-free RL~\cite{schulman2017proximal,sutton2018reinforcement} and some specific instances of model-based RL~\cite{chen2018neural, Heess15svg, hafner2019dream, hafner2019learning}. 
\end{observation}

\subsection{Online optimization and model-based RL}

\begin{observation}
Many model-based RL algorithms~\cite{fu2016one,deng2020soft,bansal2016learning} optimize the nonlinear controller cost in \autoref{eq:prob}. They limit the horizon $T$, reducing the computational load and allowing online optimization with respect to actions, e.g., via iterative Linear Quadratic Regulator (iLQR) method or MPC.
%
A drawback of such methods is that the optimization is calculated online, which is potentially both prohibitively slow and difficult to solve.
This is especially true for non-convex dynamics or dynamics with a high dimensional state space.
\end{observation}

\subsection{Differentiable programming (DP)}
\begin{observation}
If $f$ and $\pi$ are differentiable, $J^*(\pi)$ can be minimized directly via gradient descent.
\end{observation}

Such end-to-end differentiable training is also known as training with Analytical Policy Gradients (APG)~\cite{freeman2021Brax}. 
Neural ODEs~\cite{chen2018neural} have inspired APG approaches for control, and several differentiable physics simulators were developed, such as DifftTaichi~\cite{hu2019difftaichi}, DiffSim~\cite{qiao2020scalable}, or Deluca~\cite{gradu2021deluca}. 
Gradu et~al.~\cite{gradu2021deluca} also demonstrate APG applications for control tasks and include a planar quadrotor model in their experiments.
%
%
Other APG approaches were proposed for solving non-linear system identification and optimal control~\cite{jin2020pontryagin, jatavallabhula2021gradsim} which outperformed state-of-the-art control methods.
Jin et~al. propose a framework to solve complex control tasks using (deep) learning techniques \cite{jin2020pontryagin}. They find that this approach improves on state-of-the-art methods in solving non-linear system identification and optimal control.

When APG is used for trajectory tracking with a fixed-length horizon, it is closely related to a paradigm termed Backpropagation-Through-Time (BPTT), which has been extensively studied~\cite{jin2020pontryagin, Grzeszczuk98neuro,deisenroth2011pilco, Paavo18total, degrave19differentiable, clavera18modelbased}.
For instance, Bakker et~al.~\cite{bakker2003robot} use BPTT with Recurrent Neural Networks (RNN) for offline policy learning~\cite{bakker2003robot}.
Schaefer et~al.~\cite{schaefer2007recurrent} and Wierstra et~al.~\cite{wierstra2010recurrent} apply RNN- or LSTM-based versions of BPTT on the CartPole task and already demonstrated the sample-efficiency of this method.
However, BPTT has limitations. Exploding and vanishing gradients can occur when training a control policy, as discussed by Metz et~al.~\cite{metz2022Gradients} in their recent work.

\begin{figure*}[!t]
 \centering
 \begin{subfigure}{0.1\textwidth}
    \vspace{6pt}
    \includegraphics[width=\textwidth]{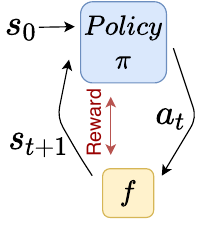}
    \caption{RL}\label{fig:modelfreerl}
    \end{subfigure}
    \hfill
    \begin{subfigure}{0.62\textwidth}
    \vspace{6pt}
    \includegraphics[width=\textwidth]{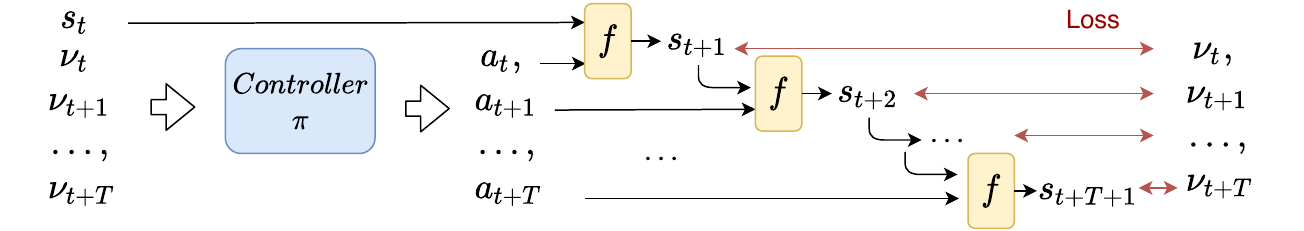}
    \caption{APG-based training (concurrent)}\label{fig:apg}
    \end{subfigure}
    \hfill
    \begin{subfigure}{0.14\textwidth}
        \vspace{9pt}
        \includegraphics[width=\textwidth]{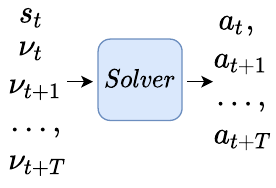}
        \caption{MPC}\label{fig:mpc}
    \end{subfigure}
 \caption{Comparison of controller training methods: Our APG controller does not require a solver, but minimizes the costs directly via gradient descent instead of maximizing a black-box reward. With the concurrent training strategy, $T$ actions $a$ are predicted simultaneously. Based on this loss, the parameters of $\pi$ are updated with backpropagation through time.
 }
     \label{fig:actions}
\end{figure*}

We apply a modified version of BPTT to control problems. Specifically, we tackle numerical training issues with curriculum learning and we explore different strategies to train with a fixed horizon. The method is explained in detail in the following, and is contrasted with MPC and RL in \autoref{fig:actions}.



%

\section{Methods}\label{sec:methods} 

In Analytic Policy Gradient learning, we use gradient descent to train the policy $\pi$ with respect to the dynamics model $f$.
Specifically, we minimize a loss function $L(\theta)$ that approximates $J^*(\pi)$ on a dataset $\mathcal{D}$:
\begin{equation}\label{eq:losspolicy}
    L(\theta) = \frac{1}{|\mathcal{D}|}\sum_{(s, \nu)\in \mathcal{D}} \sum_{t=0}^{T} C(f\big(s_t, a_t\big), \nu_{t}),
\end{equation}
where $\theta$ are the parameters of $\pi$, $\mathcal{D}$ is a set of state-reference pairs collected from interaction with $f$, and we refer to \mbox{$a_t = \pi(s_t, \nu)$} as the action at step $t$.
%
In contrast to MPC or iLQR, we can optimize $\pi$ \emph{offline} on the entire interaction data $\mathcal{D}$.
This optimization is done with stochastic optimization algorithms.
%
%
%
To favor short-term planning, we train $\pi$ in a receding-horizon fashion.
Analogous to MPC, the controller predicts $T$ actions and computes the loss on the divergence of the resulting $T$ subsequent states from the reference trajectory. The reference states in the horizon are input to the policy network at all steps, i.e. $a_t = \pi(s_t, \nu_{t}, \nu_{t+1}, \dots, \nu_{t+T})$. 

\subsection{Backpropagation over a horizon}

\begin{figure}
    \centering
    \includegraphics[width=\columnwidth]{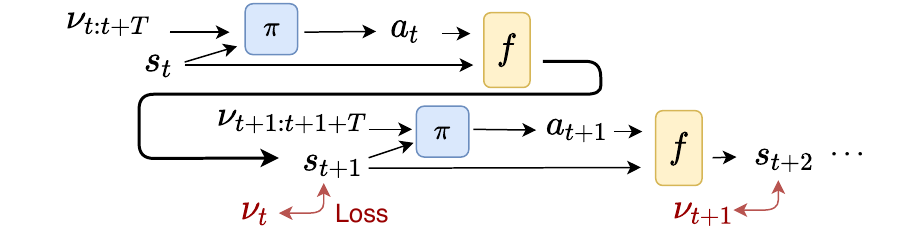}
    \caption{Training an APG controller with an autoregressive approach: Instead of predicting the next $T$ actions $a$ at once, we predict and apply one action per time step and feed the resulting state back into the network. We thereby backpropagate through the policy network multiple times.}
    \label{fig:rnn_style}
\end{figure}

At test time only one action is applied at a time.
There are, however, two possibilities to train over a horizon: 1)~At time $t$, the policy outputs $T$ actions \emph{concurrently}, and they are applied subsequently (\autoref{fig:apg}). The action at step $t$ is given by $a_t = [\pi(s_0, \nu_{0:T})]_{t}$ where $\nu_{0:T}$ denotes the $T$ reference states within the horizon and $[\cdot]_t$ is the $t$-th component of an output vector. Here, BPTT passes multiple times through $f$, but only once through $\pi$. 
2)~On the other hand, an action can be predicted in \textit{autoregressive} manner and applied in each time step, as shown \autoref{fig:rnn_style}. The action at time $t$ is $a_t = \pi(s_t, \nu_{t:t+T})$. After $T$ steps, the loss is computed with respect to the reference trajectory and the gradients are backpropagated through the whole chain of $T$ policy-network predictions and $T$ times the dynamics, resulting in a longer chain of gradients.

To the best of our knowledge, only the latter approach has been analyzed in the literature. However, it is well known that BPTT with this paradigm leads to vanishing or exploding gradients~\cite{bengio1994Learning} and was found to potentially lead to performance limitations of the policy~\cite{metz2022Gradients}. To improve training stability, we propose the concurrent controller and provide a systematic comparison.

\subsection{Curriculum learning}
\label{sec:methods-curriculum}

\begin{figure}[t]
    \centering
    \includegraphics[width=0.65\columnwidth]{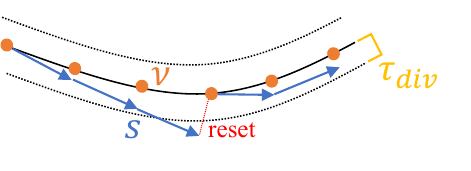}
    \caption{Curriculum learning, where we reset the state $s$ onto the trajectory $\nu$ if the error exceeds the threshold $\tau_{div}$.}
    \label{fig:curriculum}
\end{figure}

Nevertheless, large displacements of the states with respect to the reference trajectory can occur and increase the loss. Training with such a loss function is initially unstable and may end in a local optimum. 
We thus additionally propose a curriculum learning strategy to improve training, as depicted in \autoref{fig:curriculum}.
We set a threshold $\tau_{div}$ to the divergence that is applied only when collecting the dataset $\mathcal{D}$. We run the partially-trained controller to collect new training data, but whenever \mbox{$\Vert \hat{s}_{t+1} - \nu_{t+1} \Vert > \tau_{div}$} after executing action $a_t$, the agent is reset to $\nu_{t+1}$.
The threshold $\tau_{div}$ is then increased over time.
Therefore, the robot will learn to follow the reference from gradually more distant states as training proceeds.
%
%
%


%
\section{Experiments}\label{sec:experiments}
\begin{figure}
    \centering
    \vspace{6pt}
    \includegraphics[width=\columnwidth]{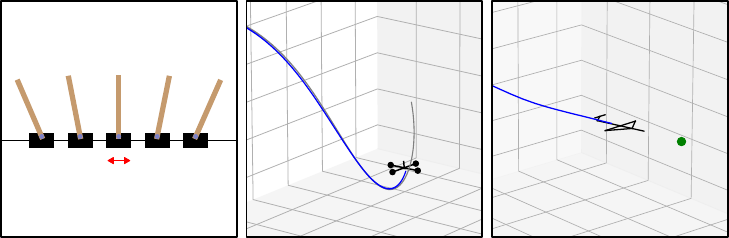}
     \caption{Experiments are conducted on three diverse dynamic systems and tasks: Balancing a pole on a cart (left), tracking a trajectory with a quadrotor (center) and passing through a target point (green) with a fixed wing drone (right).
     }
\label{fig:systems}
\end{figure}

%
%
%
We compare our approach to both classical control and reinforcement learning algorithms on three diverse tasks: CartPole balancing, trajectory tracking with a quadrotor, and flying to a 3D target with a fixed-wing aircraft (see \autoref{fig:systems}). 
We design our evaluation procedure to answer the following questions: (1)~How does our proposed APG method compare to commonly used RL and MPC algorithms? (2)~Are the measures of limiting the backpropagation chain length and introducing the curriculum improving training stability? (3)~Can the advantages of APG be leveraged to deal with high-dimensional input data (e.g. images) or to improve the sample efficiency for adaptation tasks?
%
%
%
%
\subsection{Experimental setup}\label{sec:setup}
As a low-dimensional state space task, we test APG algorithm on the CartPole balancing task where a pole must be balanced by moving a cart.
We use a model derived from first principles~\cite{barto1983neuronlike} as dynamic model $f$, and allow for continuous actions in the range $[-1,1]$ corresponding to $\pm 30$~N force on the cart.
Since the reference trajectory only consists of a single state (i.e. upright position of the pole), we linearly interpolate from the current state to the desired end state (upright and zero velocity) to generate a continuous reference $[\nu_{t+1},\dots,\nu_{t+k}]$. 

The second task consists of tracking a 3D trajectory with a quadrotor. We use the first-principles model provided by the Flightmare simulator~\cite{song2020flightmare} as the dynamic model $f$. For training and testing, we generate random polynomial trajectories with an average speed of $3~\frac{\text{m}}{\text{s}}$.
We evaluate performance based on $50$ test trajectories that were not seen during training.

Finally, for reaching a 3D target with a fixed-wing aircraft, we implement a first-order model as described by Beard~et~al.~\cite{beard2012small}. Fixed-wing aircrafts can fly significantly faster than quadrotors, but are less maneuverable and thus more challenging to control. 
In our experiment, the aircraft is initialized with a forward velocity of $11.5~\frac{\text{m}}{\text{s}}$ in the $x$-direction, and is tasked to reach a target at a distance of $x=50$ m, with $y$ and $z$ deviations from the initial position uniformly sampled from $[-5,5]$ m. 
We evaluate performance with $30$ target locations that have not been seen during training.
Similar to the CartPole balancing task, the reference trajectory only consists of a single state, i.e. the target, and we generate a continuous reference  $[\nu_{t+1},\dots,\nu_{t+k}]$ by linear interpolation from the current to the target state, however maintaining constant forward speed. 
%
Our approach works despite this approximation of the reference trajectory which may be infeasible.
%
%
We perform all experiments using PyTorch~\cite{paszke2019pytorch}; 
see the Appendix for technical details on model definition and training. We 
publish the source code on the project website to reproduce our experiments, and provide example videos of the APG trained policies on all three tasks in the supplementary material.

\subsection{Comparison to model-based RL, model-free RL, and MPC}\label{sec:experimentsrl}
We compare our best-performing APG controller to three strong baselines~(1) the model-free RL algorithm Proximal Policy Optimization (PPO)~\cite{schulman2017proximal}, 
(2)~the model-based RL algorithm termed probabilistic-ensembles-with-trajectory-sampling (PETS)~\cite{chua2018deep}, and (3)~model predictive control (MPC). 
%
The PPO baseline is available from the \texttt{stable-baselines} PyPI package~\cite{stable-baselines3}. The PETS implementation was taken from the \texttt{mbrl} toolbox~\cite{Pineda2021MBRL}. The MPC is implemented as a non-linear programming solver with \texttt{Casadi}~\cite{andersson2019casadi} and is provided in the supplementary code.
%
%
All methods have access to the same inputs, produce the same type of output, are tuned individually with grid-search on a held-out validation set, and the policies are trained with the same curriculum.
%
%
%

We train and test the methods described above on all three systems, and evaluate the performance in tracking randomly-sampled polynomial reference trajectory. For the CartPole system, we evaluate the performance based on the stability of the cart, i.e. how fast the system moves along the $x$-axis. For the quadrotor and fixed-wing tasks, performance is evaluated with the average tracking error (in m). The results of these experiments are shown in Table~\ref{tab:rltable}.

\begin{table}[t]
  \centering
  \vspace{6pt}
  \resizebox{\columnwidth}{!}{
  \begin{tabular}{@{}c|c|ccc}
  \toprule
    & 
    &  \multicolumn{3}{c}{\textbf{Learning-based control}} \\
    & \textbf{MPC} & \textbf{PPO} & \textbf{PETS}  & \textbf{Ours}\\
    \midrule
\textbf{CartPole} & & & \\
Cart velocity (in m/s)
& 0.01$\pm$0.03 & 
   \textbf{0.04 $\pm$0.04}  & 
    0.24 $\pm$0.10 & 
    0.05 $\pm$0.05 \\
    Runtime (in ms) & 4.9   &           0.28     &  2200   &    \textbf{0.16} \vspace{6pt}
    \\
    \midrule
    \textbf{Quadrotor} & & & \\
    Tracking error (in m)
    &
    0.03 $\pm$0.01 & 
    0.36 $\pm$0.10 &
    unstable & 
    \textbf{0.05 $\pm$0.01} \\
    Runtime (in ms) & 5.2   &           0.31      &  4919  &   \textbf{0.17}    \vspace{6pt}
    \\
    \midrule
    \textbf{Fixed-wing} & & & \\
    Tracking error (in m)
    &
    0.0 $\pm$0.0 & 
    0.09 $\pm$0.05  &
    0.24 $\pm$0.15 &
    \textbf{0.01 $\pm$0.01} \\
    Runtime (in ms)  & 11.4   &          0.25     &  2775  &    \textbf{0.23} \\
    \bottomrule
  \end{tabular}
  }
    \caption{
    Tracking performance and runtimes for classical and learning-based control methods. The best learning-based results are marked bold. While MPC has very low tracking error, differentiable control is the only learning-based method that achieves comparable performance. At the same time, it has significant advantages in terms of computational efficiency at inference time.
    }\label{tab:rltable}
    \vspace{-6pt}
\end{table}

From the table, we can clearly see that classical control with non-linear optimization achieves the best performance in terms of tracking error.
This is to be expected, given the availability of the model dynamics and an (approximate) reference trajectory.
The major drawback of MPC is its computational load at runtime, with a calculation time of several milliseconds for each time step.
In flying vehicles, the runtime is a major limitation and has therefore received much attention in the literature~\cite{wang2021efficient, oettershagen2014explicit, stastny2017nonlinear, hanover2021performance}.
Similarly, model-based RL methods, such as PETS, also employ online-optimization to compute the next action and thus suffer from the same problem. The runtime (and the performance) for PETS - and presumably other model-based RL methods - are worse due to the assumption of a black box model, which demands an approximation of the system dynamics with a complex learnt function. For the quadrotor tracking task, the PETS model only converged in a simplified scenario, namely tracking with lower speed ($\approx 1~\frac{\text{m}}{\text{s}}$) and only on a single reference trajectory, which still results in a tracking error of $0.18$~m. 

On the other hand, model-free RL such as PPO is fast at inference time, but leads to much larger tracking error, with the exception of the CartPole system. Overall, APG offers a way to achieve very low tracking error while maintaining the low inference runtime of learning-based methods like PPO. All times were computed on a single CPU. While better implementations could increase the performance of optimization-based methods, our approach stands out as a method that achieves both low tracking error, low sample complexity, and a low runtime.\subsection{Validation of receding-horizon training and curriculum}\label{sec:experiments_curriculum}

\subsubsection{Training with receding-horizon}
Furthermore, we provide a systematic comparison of the training strategies discussed in \autoref{sec:methods}, namely 
training with a receding horizon by predicting all actions at once (\textit{concurrent}) or sequentially (\textit{autoregressive}), see \autoref{fig:apg} and \autoref{fig:rnn_style} respectively. Furthermore, we test an autoregressive model with memory, using an LSTM\footnote{All training details for the three methods are provided in the code repository at \url{https://github.com/lis-epfl/apg_trajectory_tracking/blob/main/training_details.pdf}}. 
We evaluate these methods with varying horizon length $T$ in the task of trajectory tracking with a quadrotor, 
and evaluate performance in terms of tracking success and tracking error. Tracking success is defined as the ratio of trajectories that was followed successfully without diverging more than 5 m from $\nu$, whereas the average tracking error is only computed on the successful tracking runs.

The experiments reported in \autoref{tab:horizon} show that if the horizon is too short, i.e. below 5 steps, APG is not able to look far enough into the future to achieve a desirable trajectory tracking performance. 
However, with a long horizon, i.e. above 12 steps, the performance degrades again.
We conjecture that this is caused by the increasing length of the gradient chains, which lead to poorly tractable gradients. 
Interestingly, the autoregressive approach works well with longer horizon, but generally converges to worse tracking error. Adding memory in from of an LSTM does not improve performance for long horizon lengths (8-15) and fails when the horizon length is short (1-5). We hypothesize that this is due to the Markov property of the task.  
Given the better performance of our new approach with a horizon of $T=10$, we use this model for all other experiments.

\subsubsection{Curriculum learning}

We verify the importance of our curriculum learning for the training stability of APG by training the model with and without the curriculum 
explained in \autoref{sec:methods-curriculum}. 
During training we evaluate the control performance after each epoch by tracking 100 trajectories until the quadrotor is either too far away from the trajectory (5 m) or has reached the end of the trajectory.

\begin{figure}
    \centering
    \vspace{6pt}
    \includegraphics[width=0.8\columnwidth]{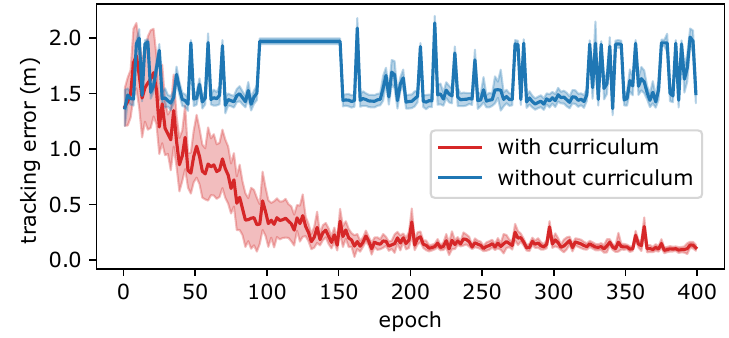}
    \caption{Tracking performance with and without curriculum learning, where the line reports the average and shaded area the standard deviation. The curriculum leads to a stable learning process.}
    \vspace{-0.2cm}
    \label{fig:curriculum_comparison}
\end{figure}

In \autoref{fig:curriculum_comparison} we report the tracking error throughout the training.
We note three things: (1)~Training with curriculum leads to a significant reduction of the tracking error, while without it does not.
(2)~There are jumps in the tracking error of the training without curriculum, which may be caused by unstable training.
(3)~There is minimal change between episodes 95 and 15 in the training without curriculum, which may have been caused by a local minimum.
%
Based on these observations, we argue that the introduction of curriculum learning stabilizes APG training and reduces the chance of ending in a local minimum.

\begin{table}[t]
  \centering
  \vspace{6pt}
  \resizebox{\columnwidth}{!}{
  \begin{tabular}{c|ccc|ccc}
    \toprule
    & \multicolumn{3}{c}{\textbf{Tracking error (in m)}} & \multicolumn{3}{c}{\textbf{Tracking success}}
    \\  
      & Con- & Auto- & Recurrent & Con- & Auto- & Recurrent \\
    T & current & regressive & (LSTM) & current & regressive & (LSTM)\\
    \midrule
    
    1 &    \--- &   \--- & \--- & 0.00 &   0.00 & 0.00\\
    3  &  2.43 $\pm$ 0.00  & \textbf{1.57} $\pm$ 0.27 & \---  & 0.01 &  0.01 & 0.00 \\
    5    &     \textbf{0.10} $\pm$ 0.05   & 0.15 $\pm$ 0.11 & \--- & 0.91 & \textbf{0.98} & 0.00\\
    8    &     \textbf{0.07} $\pm$ 0.05  &  0.09 $\pm$ 0.04 & 0.30 $\pm$ 0.09& \textbf{0.99} & 0.97 & 0.92\\
    10   &     \textbf{0.05} $\pm$ 0.01  &   0.12 $\pm$ 0.04 & 0.18 $\pm$ 0.07  & \textbf{0.99} & 0.98 & 0.97 \\ 
    12   &     \textbf{0.11} $\pm$ 0.06  &    0.17 $\pm$ 0.06 & 0.48 $\pm$ 0.13 & \textbf{0.99}& \textbf{0.99} & 0.91\\
    15   &     \--- & \textbf{0.28 }$\pm$ 0.09 & 0.61 $\pm$ 0.18  & 0.00 & \textbf{0.99} & 0.76 \\
    \bottomrule
  \end{tabular}
  }
    \caption{Table showing the influence of the control horizon length ($T$) on the control performance. If the horizon is too short, the control lacks the ability to plan ahead; if it is too long, the errors become difficult to backpropagate. 
    }
  \label{tab:horizon}
  \vspace{-0.1cm}
\end{table}

\subsection{Vision-based control}\label{sec:fromimages}

In \autoref{tab:rltable}, MPC achieved very low tracking errors in all tasks. However, besides its inferiority in terms of runtime, a shortcoming of classical control algorithms is their inability to deal with high-dimensional inputs such as images.
In contrast, here, we can demonstrate the ability of APG to learn a control policy for the CartPole task with only images as inputs. 

We use a 4-layer convolutional network with  kernel size $3$ and ReLU activations as policy. 
Instead of the 4-dimensional CartPole state, we input a sequence of four images of the CartPole in 3rd-person view (see \autoref{fig:systems}). This sequence allows the network to infer the current angular pole direction and velocity.
%
%

\autoref{tab:seqtable} shows the results. In this case, online optimization in the image space either does not converge or returns poor solutions.
%
%
In contrast, our method and the model-free baseline (PPO) both obtain very good performance. 
However, our approach can train the policy with a fraction of the sample budget (1K vs 230K samples) while obtaining a $29\%$ better performance in the target domain.
This experiment shows that our approach combines the best of both worlds. It can leverage prior knowledge about the system to increase sample efficiency and at the same time retain the flexibility of model-free methods to cope with high-dimensional inputs. 

\begin{table}[ht]
  \centering
  \vspace{6pt}
\resizebox{\columnwidth}{!}{
  \begin{tabular}{@{}lccc}
    \toprule
     & \textbf{MPC} & \textbf{PPO} & \textbf{Ours}  \\
    \cmidrule{2-4}
    \begin{tabular}{l} \textbf{Vision-based CartPole}\\ Cart velocity (in $\frac{\text{m}}{\text{s}}$) \end{tabular} 
    & unstable & 0.28 $\pm$0.23 & 0.15 $\pm$0.11 \\
    \bottomrule
  \end{tabular}
  }
    \caption{Image-based control and identification for the task of CartPole balancing. 
    Our approach outperforms the baselines in terms of tracking performance.}
  \label{tab:seqtable}
  \vspace{-6pt}
\end{table}
%
%
\subsection{Exploring the sample efficiency in adaptation tasks}

In contrast to other learning-based methods, APG is highly sample-efficient. This is particularly relevant for adaptation scenarios, where we need to learn from real-world examples. For example, consider a scenario where the dynamics of a quadrotor in simulation (source domain $f$) \emph{differ} from the dynamics in reality (target domain $f^*$). The sample size to adapt to the new dynamics should be kept as low as possible. We evaluate the adaptation performance of APG in a benchmark problem for flying vehicles, which is to estimate and counteract a linear translational drag acting on the platform~\cite{faessler2017differential,franchi2018full}. This force is modeled as \mbox{$\dot{v}\longleftarrow \dot{v} - r\cdot v$} where $v$ is the three dimensional velocity and $r=0.3~\frac{\text{1}}{\text{s}}$ the drag factor.

Assuming that the reason of the change in dynamics (i.e. the drag) is unknown, we propose to train a neural network $\Delta_{\phi}$ that acts as a residual on top of the original dynamics model $f$, essentially learning the difference between $f$ and target $f^*$. The residual is trained on a small dataset $\mathcal{D}_\text{dyn}$ of state-action-state triples:
\begin{equation}\label{eq:target_data}
    \mathcal{D}_{\text{dyn}} = \{(s_{t}, a_t, s_{t+1}^*)\ |\ t\in[1..B],\  s^*_{t+1} = f^*(s_t, a_t)\},
\end{equation}

$\mathcal{D}_{\text{dyn}}$ is used to minimize the difference of source and target domain by training a state-residual network $\Delta_\phi$ with the following loss function:
\begin{gather}\label{eq:mindelta}
    L_{\text{dyn}}(\phi) = \sum_{\mathcal{D}_\text{dyn}} \Vert f(s_t, a_t) + \Delta_\phi(s_t, a_t) - s^*_{t+1} \Vert,
\end{gather}
where $\phi$ are the parameters of the network $\Delta_\phi$ that accounts for effects that are not modeled in $f$. The network $\Delta_\phi$ can be conditioned on any kind of information, for example images, and can account for arbitrary unmodeled dynamics effects in the target domain.

\begin{figure}
    \centering
    \includegraphics[width=\columnwidth]{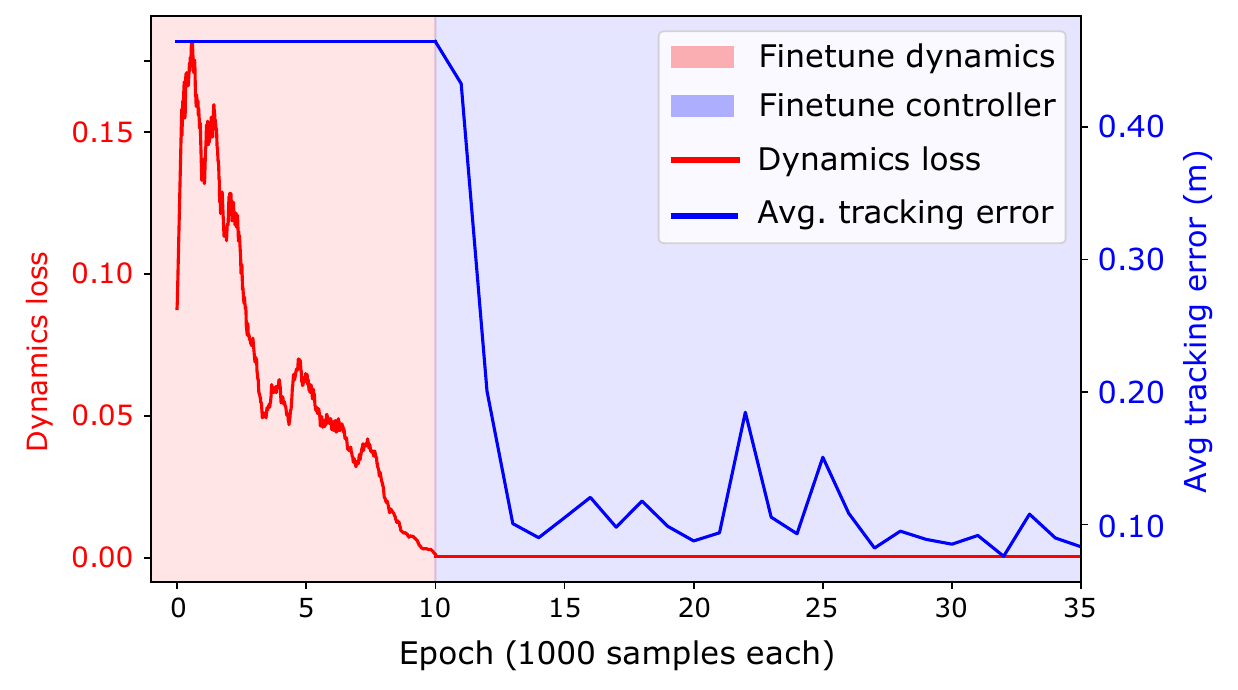}
    \caption{Fine-tuning an APG controller for adaptation. A residual dynamics model is trained (red) and the controller is fine-tuned (blue). It converges with few samples.}
    \vspace{-6pt}
    \label{fig:dyn_con}
\end{figure}

An example is given in \autoref{fig:dyn_con}. After the dynamics $\Delta_{\phi}$ are trained with samples from $f^*$ (red), the pre-trained APG controller is fine-tuned and converges with few samples. \autoref{tab:adaptation} provides the results upon fine-tuning pre-trained policies on dynamics with velocity drag. Our method, i.e. training a residual on a differentiable dynamics model and fine-tuning a differentiable learnt policy, requires orders of magnitudes fewer samples than RL methods that assume a black-box dynamics model.

\begin{table}[!htb]
  \centering
  \vspace{6pt}
  \resizebox{\columnwidth}{!}{
  \begin{tabular}{l|lccc}
    \toprule
    \textbf{Task (\#samples)} & \textbf{PPO} & \textbf{PETS}  & \textbf{Ours}\\ 
    \midrule
    
    \textbf{Quadrotor} & & & & \\
    Source domain $f$ &  
    0.36 $\pm$ 0.10 &
    unstable & 
    \textbf{0.05 $\pm$ 0.01} \\
    
    Zero-shot to $f^*$ &  
    \textbf{0.42 $\pm$ 0.09} &
    unstable &
    0.56 $\pm$ 0.07  \\
    
    Few-shot to $f^*$ (1K) & 
    0.42 $\pm$ 0.09 & 
    unstable & 
    \textbf{0.07 $\pm$ 0.01} \\
    
    Many-shot to $f^*$ (150K) &
    0.40 $\pm$ 0.12  &
    unstable &
    -- \\
    \midrule
    %
    \textbf{Fixed-wing} & & & & \\

    Source domain $f$ & 
    0.09 $\pm$ 0.05  &
    0.24 $\pm$ 0.15 &
    \textbf{0.01 $\pm$ 0.01} \\ 
    
    Zero-shot to $f^*$ &  
    0.43 $\pm$ 1.45 &
    8.39 $\pm$ 10.15 &
    \textbf{0.07 $\pm$ 0.10} \\
    
    Few-shot to $f^*$ (2K) & 
    0.34 $\pm$ 0.11 & 
    5.86 $\pm$ 5.86 & 
    \textbf{0.02 $\pm$ 0.03} \\
    
    Few-shot to $f^*$ (5K) &
    0.17  $\pm$ 0.07 &
    0.27 $\pm$ 0.14 &
    -- \\
    
    Many-shot to $f^*$ (150K) &
    0.09 $\pm$ 0.06  &
    -- &
    -- \\
    \bottomrule
  \end{tabular}}
    \caption{Adaptation to velocity drag. Our APG method converges to the initial performance with few samples, whereas other RL approaches require significantly more training.}
  \label{tab:adaptation}
  \vspace{-6pt}

\end{table}

\section{Conclusion}\label{discussion}

APG algorithms have the potential to rely on little training data and allow fast computation during runtime, bridging the gap between RL and MPC.
We proposed a receding-horizon implementation of APG with a curriculum learning scheme that improves training stability and enables the application of APG on more complex and longer-horizon problems.
Through controlled experiments on a CartPole, a quadrotor, and a fixed-wing task, we showed that our approach outperforms both commonly used model-free and model-based RL algorithms in terms of tracking error.
We showed that the tracking performance is comparable to MPC, while reducing the computation time on deployment by more than an order of magnitude.
%
%
%
Finally, we highlight the advantages of learning-based control by applying our APG controller on a vision-based control task, and demonstrate the sample efficiency of APG when adaptating to new environments.

We note that our problem formulation as a fixed-horizon tracking task implies two limitations.
Firstly, it is only demonstrated on control tasks with a relatively short horizon (up to $\SI{5}{\second}$ in the future).
A long horizon is in principle possible but may lead to training instability (e.g. exploding gradients). 
%
%
Secondly, we tested our framework on settings with (a) known dynamics and (b) given reference trajectory only.
Future research on real robots will therefore be necessary and may lead to additional insights. 
However, this paper's evaluation of common robotics tasks in simulation, provides initial evidence that the presented approach works with highly non-realistic references (e.g. linear trajectory in fixed-wing flight), and with learned dynamics.
We thus consider gradient-based policies, in particular, combined with curriculum learning schemes, promising for various control tasks. It offers a good trade-off between performance, efficiency, and flexibility, and should receive more attention in the control field.



\addtolength{\textheight}{-0cm}   




\section*{ACKNOWLEDGMENTS}

We thank Elia Kaufmann, René Ranftl, and Yunlong Song for the fruitful discussion throughout the project.


\bibliographystyle{IEEEtran}
\bibliography{references}

\begin{thebibliography}{10}
\providecommand{\url}[1]{#1}
\csname url@samestyle\endcsname
\providecommand{\newblock}{\relax}
\providecommand{\bibinfo}[2]{#2}
\providecommand{\BIBentrySTDinterwordspacing}{\spaceskip=0pt\relax}
\providecommand{\BIBentryALTinterwordstretchfactor}{4}
\providecommand{\BIBentryALTinterwordspacing}{\spaceskip=\fontdimen2\font plus
\BIBentryALTinterwordstretchfactor\fontdimen3\font minus
  \fontdimen4\font\relax}
\providecommand{\BIBforeignlanguage}[2]{{%
\expandafter\ifx\csname l@#1\endcsname\relax
\typeout{** WARNING: IEEEtran.bst: No hyphenation pattern has been}%
\typeout{** loaded for the language `#1'. Using the pattern for}%
\typeout{** the default language instead.}%
\else
\language=\csname l@#1\endcsname
\fi
#2}}
\providecommand{\BIBdecl}{\relax}
\BIBdecl

\bibitem{li2019fast}
Y.~Li, H.~Li, Z.~Li, H.~Fang, A.~K. Sanyal, Y.~Wang, and Q.~Qiu, ``Fast and
  accurate trajectory tracking for unmanned aerial vehicles based on deep
  reinforcement learning,'' pp. 1--9, 2019.

\bibitem{yu2017deep}
R.~Yu, Z.~Shi, C.~Huang, T.~Li, and Q.~Ma, ``Deep reinforcement learning based
  optimal trajectory tracking control of autonomous underwater vehicle,'' pp.
  4958--4965, 2017.

\bibitem{salzmann2022neural}
T.~Salzmann, E.~Kaufmann, M.~Pavone, D.~Scaramuzza, and M.~Ryll, ``Neural-mpc:
  Deep learning model predictive control for quadrotors and agile robotic
  platforms,'' \emph{arXiv preprint arXiv:2203.07747}, 2022.

\bibitem{wang2021efficient}
D.~Wang, Q.~Pan, Y.~Shi, J.~Hu \emph{et~al.}, ``Efficient nonlinear model
  predictive control for quadrotor trajectory tracking: Algorithms and
  experiment,'' \emph{IEEE Transactions on Cybernetics}, vol.~51, no.~10, pp.
  5057--5068, 2021.

\bibitem{abdolhosseini2013efficient}
M.~Abdolhosseini, Y.~M. Zhang, and C.~A. Rabbath, ``An efficient model
  predictive control scheme for an unmanned quadrotor helicopter,''
  \emph{Journal of intelligent \& robotic systems}, vol.~70, no.~1, pp. 27--38,
  2013.

\bibitem{song2021autonomous}
Y.~Song, M.~Steinweg, E.~Kaufmann, and D.~Scaramuzza, ``Autonomous drone racing
  with deep reinforcement learning,'' in \emph{2021 IEEE/RSJ International
  Conference on Intelligent Robots and Systems (IROS)}.\hskip 1em plus 0.5em
  minus 0.4em\relax IEEE, 2021, pp. 1205--1212.

\bibitem{freeman2021Brax}
C.~D. Freeman, E.~Frey, A.~Raichuk, S.~Girgin, I.~Mordatch, and O.~Bachem,
  ``Brax-a differentiable physics engine for large scale rigid body
  simulation,'' in \emph{Thirty-fifth Conference on Neural Information
  Processing Systems Datasets and Benchmarks Track (Round 1)}, 2021.

\bibitem{hu2019difftaichi}
Y.~Hu, L.~Anderson, T.-M. Li, Q.~Sun, N.~Carr, J.~Ragan-Kelley, and F.~Durand,
  ``Difftaichi: Differentiable programming for physical simulation,'' in
  \emph{International Conference on Learning Representations}, 2020.

\bibitem{de2018end}
F.~de~Avila Belbute-Peres, K.~Smith, K.~Allen, J.~Tenenbaum, and J.~Z. Kolter,
  ``End-to-end differentiable physics for learning and control,''
  \emph{Advances in neural information processing systems}, vol.~31, 2018.

\bibitem{innes2019differentiable}
M.~Innes, A.~Edelman, K.~Fischer, C.~Rackauckas, E.~Saba, V.~B. Shah, and
  W.~Tebbutt, ``A differentiable programming system to bridge machine learning
  and scientific computing,'' \emph{arXiv preprint arXiv:1907.07587}, 2019.

\bibitem{qiao2020scalable}
Y.-L. Qiao, J.~Liang, V.~Koltun, and M.~C. Lin, ``Scalable differentiable
  physics for learning and control,'' \emph{International Conference on Machine
  Learning}, 2020.

\bibitem{bengio1994Learning}
Y.~Bengio, P.~Simard, and P.~Frasconi, ``Learning long-term dependencies with
  gradient descent is difficult,'' \emph{IEEE Transactions on Neural Networks},
  vol.~5, no.~2, 1994.

\bibitem{gillen2022Leveraging}
S.~Gillen and K.~Byl, ``Leveraging reward gradients for reinforcement learning
  in differentiable physics simulations,'' \emph{arXiv preprint
  arXiv:2203.02857}, 2022.

\bibitem{holl2020Learning}
P.~Holl, N.~Thuerey, and V.~Koltun, ``Learning to control pdes with
  differentiable physics,'' in \emph{International Conference on Learning
  Representations}, 2020.

\bibitem{oettershagen2014explicit}
P.~Oettershagen, A.~Melzer, S.~Leutenegger, K.~Alexis, and R.~Siegwart,
  ``Explicit model predictive control and l 1-navigation strategies for
  fixed-wing uav path tracking,'' \emph{IEEE 22nd Mediterranean Conference on
  Control and Automation}, pp. 1159--1165, 2014.

\bibitem{borrelli2017predictive}
F.~Borrelli, A.~Bemporad, and M.~Morari, \emph{Predictive control for linear
  and hybrid systems}.\hskip 1em plus 0.5em minus 0.4em\relax Cambridge
  University Press, 2017.

\bibitem{schulman2017proximal}
J.~Schulman, F.~Wolski, P.~Dhariwal, A.~Radford, and O.~Klimov, ``Proximal
  policy optimization algorithms,'' \emph{arXiv preprint arXiv:1707.06347},
  2017.

\bibitem{sutton2018reinforcement}
R.~S. Sutton and A.~G. Barto, \emph{Reinforcement learning: An
  introduction}.\hskip 1em plus 0.5em minus 0.4em\relax MIT press, 2018.

\bibitem{chen2018neural}
R.~T. Chen, Y.~Rubanova, J.~Bettencourt, and D.~Duvenaud, ``Neural ordinary
  differential equations,'' \emph{NeurIPS}, 2018.

\bibitem{Heess15svg}
N.~Heess, G.~Wayne, D.~Silver, T.~Lillicrap, T.~Erez, and Y.~Tassa, ``Learning
  continuous control policies by stochastic value gradients,'' \emph{Advances
  in Neural Information Processing Systems}, vol.~28, 2015.

\bibitem{hafner2019dream}
D.~Hafner, T.~Lillicrap, J.~Ba, and M.~Norouzi, ``Dream to control: Learning
  behaviors by latent imagination,'' in \emph{International Conference on
  Learning Representations}, 2019.

\bibitem{hafner2019learning}
D.~Hafner, T.~Lillicrap, I.~Fischer, R.~Villegas, D.~Ha, H.~Lee, and
  J.~Davidson, ``Learning latent dynamics for planning from pixels,'' in
  \emph{International Conference on Machine Learning}.\hskip 1em plus 0.5em
  minus 0.4em\relax PMLR, 2019, pp. 2555--2565.

\bibitem{fu2016one}
J.~Fu, S.~Levine, and P.~Abbeel, ``One-shot learning of manipulation skills
  with online dynamics adaptation and neural network priors,'' in \emph{2016
  IEEE/RSJ International Conference on Intelligent Robots and Systems
  (IROS)}.\hskip 1em plus 0.5em minus 0.4em\relax IEEE, 2016, pp. 4019--4026.

\bibitem{deng2020soft}
Y.~Deng, Y.~Zhang, X.~He, S.~Yang, Y.~Tong, M.~Zhang, D.~DiPietro, and B.~Zhu,
  ``Soft multicopter control using neural dynamics identification,'' in
  \emph{Conference on Robot Learning}.\hskip 1em plus 0.5em minus 0.4em\relax
  PMLR, 2021, pp. 1773--1782.

\bibitem{bansal2016learning}
S.~Bansal, A.~K. Akametalu, F.~J. Jiang, F.~Laine, and C.~J. Tomlin, ``Learning
  quadrotor dynamics using neural network for flight control,'' in \emph{2016
  IEEE 55th Conference on Decision and Control (CDC)}.\hskip 1em plus 0.5em
  minus 0.4em\relax IEEE, 2016, pp. 4653--4660.

\bibitem{gradu2021deluca}
P.~Gradu, J.~Hallman, D.~Suo, A.~Yu, N.~Agarwal, U.~Ghai, K.~Singh, C.~Zhang,
  A.~Majumdar, and E.~Hazan, ``Deluca--a differentiable control library:
  Environments, methods, and benchmarking,'' \emph{arXiv preprint
  arXiv:2102.09968}, 2021.

\bibitem{jin2020pontryagin}
W.~Jin, Z.~Wang, Z.~Yang, and S.~Mou, ``Pontryagin differentiable programming:
  An end-to-end learning and control framework,'' \emph{Advances in Neural
  Information Processing Systems}, vol.~33, pp. 7979--7992, 2020.

\bibitem{jatavallabhula2021gradsim}
J.~K. Murthy, M.~Macklin, F.~Golemo, V.~Voleti, L.~Petrini, M.~Weiss,
  B.~Considine, J.~Parent-L{\'e}vesque, K.~Xie, K.~Erleben \emph{et~al.},
  ``gradsim: Differentiable simulation for system identification and visuomotor
  control,'' in \emph{International Conference on Learning Representations},
  2021.

\bibitem{Grzeszczuk98neuro}
R.~Grzeszczuk, D.~Terzopoulos, and G.~Hinton, ``Neuroanimator: Fast neural
  network emulation and control of physics-based models,'' in \emph{Proceedings
  of the 25th annual conference on Computer graphics and interactive
  techniques}, 1998, pp. 9--20.

\bibitem{deisenroth2011pilco}
M.~Deisenroth and C.~E. Rasmussen, ``Pilco: A model-based and data-efficient
  approach to policy search,'' in \emph{Proceedings of the International
  Conference on Machine Learning}, 2011, pp. 465--472.

\bibitem{Paavo18total}
P.~Parmas, ``Total stochastic gradient algorithms and applications in
  reinforcement learning,'' \emph{Advances in Neural Information Processing
  Systems}, vol.~31, 2018.

\bibitem{degrave19differentiable}
J.~Degrave, M.~Hermans, J.~Dambre \emph{et~al.}, ``A differentiable physics
  engine for deep learning in robotics,'' \emph{Frontiers in neurorobotics},
  p.~6, 2019.

\bibitem{clavera18modelbased}
I.~Clavera, J.~Rothfuss, J.~Schulman, Y.~Fujita, T.~Asfour, and P.~Abbeel,
  ``Model-based reinforcement learning via meta-policy optimization,'' in
  \emph{Conference on Robot Learning}, 2018, pp. 617--629.

\bibitem{bakker2003robot}
B.~Bakker, V.~Zhumatiy, G.~Gruener, and J.~Schmidhuber, ``A robot that
  reinforcement-learns to identify and memorize important previous
  observations,'' vol.~1, pp. 430--435, 2003.

\bibitem{schaefer2007recurrent}
A.~M. Schaefer, S.~Udluft, and H.-G. Zimmermann, ``A recurrent control neural
  network for data efficient reinforcement learning,'' in \emph{2007 IEEE
  International Symposium on Approximate Dynamic Programming and Reinforcement
  Learning}.\hskip 1em plus 0.5em minus 0.4em\relax IEEE, 2007, pp. 151--157.

\bibitem{wierstra2010recurrent}
D.~Wierstra, A.~F{\"o}rster, J.~Peters, and J.~Schmidhuber, ``Recurrent policy
  gradients,'' \emph{Logic Journal of the IGPL}, vol.~18, no.~5, pp. 620--634,
  2010.

\bibitem{metz2022Gradients}
L.~Metz, C.~D. Freeman, S.~S. Schoenholz, and T.~Kachman, ``Gradients are not
  all you need,'' \emph{arXiv preprint arXiv:2111.05803}, 2021.

\bibitem{barto1983neuronlike}
A.~G. Barto, R.~S. Sutton, and C.~W. Anderson, ``Neuronlike adaptive elements
  that can solve difficult learning control problems,'' \emph{IEEE transactions
  on systems, man, and cybernetics}, no.~5, pp. 834--846, 1983.

\bibitem{song2020flightmare}
Y.~Song, S.~Naji, E.~Kaufmann, A.~Loquercio, and D.~Scaramuzza, ``Flightmare: A
  flexible quadrotor simulator,'' in \emph{Conference on Robot Learning}.\hskip
  1em plus 0.5em minus 0.4em\relax PMLR, 2021, pp. 1147--1157.

\bibitem{beard2012small}
R.~W. Beard and T.~W. McLain, \emph{Small unmanned aircraft: Theory and
  practice}.\hskip 1em plus 0.5em minus 0.4em\relax Princeton university press,
  2012, pp. 16, 44ff.

\bibitem{paszke2019pytorch}
A.~Paszke, S.~Gross, F.~Massa, A.~Lerer, J.~Bradbury, G.~Chanan, T.~Killeen,
  Z.~Lin, N.~Gimelshein, L.~Antiga \emph{et~al.}, ``Pytorch: An imperative
  style, high-performance deep learning library,'' \emph{Advances in Neural
  Information Processing Systems}, vol.~32, pp. 8026--8037, 2019.

\bibitem{chua2018deep}
K.~Chua, R.~Calandra, R.~McAllister, and S.~Levine, ``Deep reinforcement
  learning in a handful of trials using probabilistic dynamics models,''
  \emph{Advances in Neural Information Processing Systems}, vol.~31, 2018.

\bibitem{stable-baselines3}
A.~Raffin, A.~Hill, M.~Ernestus, A.~Gleave, A.~Kanervisto, and N.~Dormann,
  ``Stable baselines3,'' \url{https://github.com/DLR-RM/stable-baselines3},
  2019.

\bibitem{Pineda2021MBRL}
L.~Pineda, B.~Amos, A.~Zhang, N.~O. Lambert, and R.~Calandra, ``Mbrl-lib: A
  modular library for model-based reinforcement learning,'' \emph{arXiv
  preprint arXiv:2104.10159}, 2021.

\bibitem{andersson2019casadi}
J.~A. Andersson, J.~Gillis, G.~Horn, J.~B. Rawlings, and M.~Diehl, ``Casadi: a
  software framework for nonlinear optimization and optimal control,''
  \emph{Mathematical Programming Computation}, vol.~11, no.~1, pp. 1--36, 2019.

\bibitem{stastny2017nonlinear}
T.~J. Stastny, A.~Dash, and R.~Siegwart, ``Nonlinear mpc for fixed-wing uav
  trajectory tracking: Implementation and flight experiments,'' in \emph{AIAA
  guidance, navigation, and control conference}, 2017, p. 1512.

\bibitem{hanover2021performance}
D.~Hanover, P.~Foehn, S.~Sun, E.~Kaufmann, and D.~Scaramuzza, ``Performance,
  precision, and payloads: Adaptive nonlinear mpc for quadrotors,'' \emph{IEEE
  Robotics and Automation Letters}, vol.~7, no.~2, pp. 690--697, 2021.

\bibitem{faessler2017differential}
M.~Faessler, A.~Franchi, and D.~Scaramuzza, ``Differential flatness of
  quadrotor dynamics subject to rotor drag for accurate tracking of high-speed
  trajectories,'' \emph{IEEE Robotics and Automation Letters}, vol.~3, no.~2,
  pp. 620--626, 2017.

\bibitem{franchi2018full}
A.~Franchi, R.~Carli, D.~Bicego, and M.~Ryll, ``Full-pose tracking control for
  aerial robotic systems with laterally bounded input force,'' \emph{IEEE
  Transactions on Robotics}, vol.~34, no.~2, pp. 534--541, 2018.

\bibitem{zhou2019continuity}
Y.~Zhou, C.~Barnes, J.~Lu, J.~Yang, and H.~Li, ``On the continuity of rotation
  representations in neural networks,'' in \emph{Proceedings of the IEEE/CVF
  Conference on Computer Vision and Pattern Recognition}, 2019, pp. 5745--5753.

\bibitem{waldock2018learning}
A.~Waldock, C.~Greatwood, F.~Salama, and T.~Richardson, ``Learning to perform a
  perched landing on the ground using deep reinforcement learning,''
  \emph{Journal of Intelligent \& Robotic Systems}, vol.~92, no.~3, pp.
  685--704, 2018.

\end{thebibliography}

\appendices
\section*{APPENDIX \\ Hyperparameters and training details}\label{appendix_details}
We tested our method on three dynamic systems that are visualized in \autoref{fig:systems}: the CartPole balancing task, trajectory tracking with a quadrotor and point-goal navigation with a fixed-wing aircraft. 
Our framework is implemented in Pytorch, enabling the use of autograd for backpropagation through time. All source code is attached, and the parameters can be found as part of the source code in the folder \texttt{configs}. Example videos for all three applications are also attached in the supplementary material. In the following we provide  details on training and evaluation.

\subsection{CartPole}\label{appendix_cartpole}
In the CartPole problem, a pole should be balanced on a cart by pushing the cart left and right, while maintaining low cart velocity. The time step is again set to $0.05$~s. The state of the system can be described by position $x$ and velocity $\dot{x}$ of the cart, as well as angle $\alpha$ and angular velocity $\dot{\alpha}$ of the pole. The ''reference trajectory'' is defined only by the target state, which is the upright position of the pole, $\alpha=0$, $\dot{\alpha}=0$ and $\dot{x}=0$.

\mypara{Policy network}
Since the reference state is constant, it is sufficient for the policy to observe the state at each time step. Here, we input the raw state without normalization. It is passed through a five-layer MLP with 32, 64, 64, 32 and 10 neurons respectively. All layers including the output layer use tanh activation, in order to scale the actions to values between $-1$ (corresponding to $30$~N force to the left) and $1$ (force of $30$~N pushing the cart to the right). As for the quadrotor and fixed-wing drone, the next $10$ (1-dimensional) actions are predicted.

\mypara{Loss function} To compute the loss with respect to a reference trajectory, we interpolate between the current state and the target state. Note that the intermediate states are often infeasible to reach; for example an increase of velocity might be required to reduce the pole angle. The interpolation where both velocity and angle decrease is thus not realistic. As before, a weighted MSE between the reference states and the actual states is computed, where the angle difference is weighted with a factor of $10$, the cart velocity with factor $3$, and the angular velocity with factor $1$. The loss is minimized with SGD optimizer with learning rate $10^{-7}$.

\subsection{Quadrotor}

Our implementation of a quadrotor environment is loosely based on the implementation provided at \url{https://github.com/ngc92/quadgym} (MIT license). However, our model of the quadrotor is a Pytorch implementation of the equations in the Flightmare simulator~\cite{song2020flightmare}. We also provide an interface to test our models in Flightmare, and a video of our model controlling a quadrotor in Flightmare is attached in the supplementary material. In all our experiments, we set the time between the discrete time steps to $0.1$~s. \\
Furthermore, for training and testing we generate $10000$ random polynomials of $10s$ length, where all trajectories are guaranteed to be feasible to track with the used platform. From this dataset, $1000$ trajectories are left out as a test set to ensure that the policy generalizes to any given reference. The maximum desired velocity on such a reference trajectory is $3-5~\frac{m}{s}$, whereas the average velocity is $1-2~\frac{m}{s}$.

\mypara{Policy network} At each time step, the current state and the next reference states are given as input to the network. As the current state, we input the velocity in the world frame and in the body frame, the first two columns of the rotation matrix to describe the attitude (as recommended in~\cite{zhou2019continuity}), and the angular velocity. For the reference, we input the next 10 desired positions \textit{relative} to the current drone position, as well as the next 10 desired velocities (in the world frame).\\
The state is first passed through a linear layer with 64 neurons (tanh activation), while the reference is processed with a 1D convolutional layer (20 filters, kernel size 3) to extract time-series features of the reference. The outputs are concatenated and fed through three layers of 64 neurons each with tanh activation. The output layer has 40 neurons to output 10 four-dimensional actions. Here, an action corresponds to the total thrust $T$ and the desired body rates $\omega_{des}$ to be applied to the system. The network outputs are first normalized with a sigmoid activation and then rescaled to output a thrust between $2.21$~N and $17.31$~N (such that an output of $0.5$ corresponds to $9.81$~N) and body rates between $-0.5$ and $0.5$.

\mypara{Loss function} The loss (equation~\ref{eq:losspolicy}) is a weighted MSE between the reached states and the reference states. The weights are aligned to the ones used for the optimization-based MPC, namely a weight of $10$ for the position loss, $1$ for the velocity, $5$ to regularize the predicted thrust command, and $0.1$ for the predicted body rates as well as the actual angular velocity. Formally, these weights yield the following loss:
\begin{multline}
     L = \sum_{k=1}^{10} 10\cdot (x_{t+k, \pi} - x_{t+k, \nu})^2 +  (\dot{x}_{t+k,\pi} - \dot{x}_{t+k, \nu})^2 +\\ 5 \cdot (T_{k} - 0.5)^2 + 0.1 \cdot (\omega_{k, des} - 0.5)^2 + 0.1 \cdot \omega_{t+k}
\end{multline}
where $x_{t}$ is the position at time step $t$, $\dot{x}_t$ is the velocity, and the subscript $\pi$ indicates the states reached with the policy while the subscript $\nu$ denotes the positions and velocities of the reference. $T$ and $\omega_{des}$ correspond to the action after sigmoid activation but before rescaling (such that values lie between $0$ and $1$), and $\omega_{t}$ is the actual angular velocity of the system at each state. 
The loss is minimized with the Pytorch SGD optimizer with a learning rate of $10^{-5}$ and momentum of $0.9$.

\mypara{Curriculum learning} As explained in \autoref{sec:methods}, we use a curriculum learning strategy with a threshold $\tau_{div}$ on the allowed divergence from the reference trajectory. We set $\tau_{div}=0.1$~m initially and increase it by $0.05$~m every $5$ epochs, until reaching $2~$m. Additionally, we start by training on slower reference trajectories (half the speed). Once the quadrotor is stable and tracks the full reference without hitting $\tau_{div}$, the speed is increased to $75\%$ of the desired speed and $\tau_{div}$ is reset to $0.1$~m. This is repeated to train at the full speed in the third iteration.

\subsection{Fixed-wing drone}

We implemented a realistic model of the dynamics based on the equations and parameters described in \cite{beard2012small} and \cite{waldock2018learning}. In our discrete-time formulation, we set the time step to $0.05$~s.

\mypara{Reference trajectory} In contrast to the reference trajectories for the quadrotor, the reference for the fixed-wing aircraft is only given implicitly with the target position. As an approximate reference, we train the policy to follow the linear trajectory towards the target point. In the following, the term ''linear reference'' will be used to refer to the straight line from the current position to the target point. 
At each time step, we compute the next 10 desired states as the positions on the linear reference while assuming constant velocity.

\mypara{Policy network}
Similar to the quadrotor, the state and reference are pre-processed before being input to the network policy. The state is normalized by subtracting the mean and dividing by the standard deviation per variable in the state (mean and standard deviation are computed over a dataset of states encountered with a random policy). This normalized state together with the relative position of the 10-th desired state on the reference are passed to the policy network as inputs. Using all 10 reference states as input is redundant since they are only equally-distant positions on a line.\\
The state and the reference are each separately fed through a linear layer with 64 neurons and then concatenated. The feed-forward network then corresponds to the one used for the quadrotor training (three further layers with 64 neurons each and an output layer with 40 neurons). The output actions are also normalized with sigmoid activations and then scaled to represent thrust $T\in[0,7]$~N, elevator angle $a_1\in [-20, 20]~{}^\circ$, aileron angle $a_2\in [-2.5, 2.5]~{}^\circ$ and rudder angle $a_3\in [-20, 20]~{}^\circ$.

\mypara{Loss function}
As for the quadrotor, we align the loss function to the cost function of the online optimization model predictive control in the MPC baseline. The MSE between the reached positions and the target positions on the linear reference is minimized while regularizing the action, formally
\begin{multline}
L = \sum_{k=1}^{10} 10 \cdot (x_{t+k, \pi} - x_{t+k, \nu})^2 + 0.1 \cdot ((a_{k, 1} - 0.5)^2 +\\ (a_{k, 2} - 0.5)^2 + (a_{k, 3} - 0.5)^2)\ ,
\end{multline}
where $t$ is the current time step and $x_{t+k, \pi}$ is the position of the aircraft after executing the k-th action $(T_k, a_{k,1}, a_{k,2}, a_{k,3})$, and $x_{t+k, \nu}$ is the corresponding reference state. 
The loss is minimized with an SGD optimizer with a learning rate of $10^{-4}$ and momentum of $0.9$. 

Finally, the curriculum is initialized to allow divergence of $4m$ from the linear reference and increased by $0.5$~m every epoch until reaching $20$~m. Note that in contrast to the quadrotor, the model converges in a few epochs.

\end{document}